\documentclass[letterpaper]{article} 
\usepackage{aaai2026}  
\usepackage{times}  
\usepackage{helvet}  
\usepackage{courier}  
\usepackage[hyphens]{url}  
\usepackage{graphicx} 
\usepackage{natbib}  
\usepackage{caption} 
\usepackage{subfigure}
\usepackage{amsmath}
\usepackage{amssymb}
\usepackage{amsthm}
\usepackage{multirow}
\usepackage{booktabs}
\usepackage[table]{xcolor}

\usepackage{algorithm}
\usepackage{algorithmic}

\usepackage{newfloat}

\usepackage{listings}
\DeclareCaptionStyle{ruled}{labelfont=normalfont,labelsep=colon,strut=off} 
\floatstyle{ruled}
\newfloat{listing}{tb}{lst}{}
\floatname{listing}{Listing}
\title{Communication-Efficient Federated Recommendation with PEFT Embeddings}
\title{Communication-Efficient Federated Recommendation with \\Parameter-Efficient Fine-Tuning Embeddings}
\title{Plug-and-Play Parameter-Efficient Fine-Tuning of Embeddings for \\Communication-Efficient Federated Recommendation}
\title{Plug-and-Play Parameter-Efficient Tuning of Embeddings for \\Federated Recommendation}

\author{
Haochen Yuan\textsuperscript{\rm 1},
Yang Zhang\textsuperscript{\rm 2}\equalcontrib,
Xiang He\textsuperscript{\rm 1},
Quan Z. Sheng\textsuperscript{\rm 3},
Zhongjie Wang\textsuperscript{\rm 1}\equalcontrib
}
\affiliations{
\textsuperscript{\rm 1}Faculty of Computing, Harbin Institute of Technology\\
\textsuperscript{\rm 2}Anuradha and Vikas Sinha Department of Data Science, University of North Texas\\
\textsuperscript{\rm 3}School of Computing, Macquarie University\\
yhc@stu.hit.edu.cn,
yang.zhang@unt.edu,
hexiang@hit.edu.cn,
michael.sheng@mq.edu.au,
rainy@hit.edu.cn
}

\usepackage{bibentry}

\begin{document}

\maketitle

\begin{abstract}
With the rise of cloud-edge collaboration, recommendation services are increasingly trained in distributed environments. Federated Recommendation (FR) enables such multi-end collaborative training while preserving privacy by sharing model parameters instead of raw data.
However, the large number of parameters, primarily due to the massive item embeddings, significantly hampers communication efficiency. While existing studies mainly focus on improving the efficiency of FR models, they largely overlook the issue of embedding parameter overhead. To address this gap, we propose a FR training framework with Parameter-Efficient Fine-Tuning (PEFT) based embedding designed to reduce the volume of embedding parameters that need to be transmitted. Our approach offers a lightweight, plugin-style solution that can be seamlessly integrated into existing FR methods. In addition to incorporating common PEFT techniques such as LoRA and Hash-based encoding, we explore the use of Residual Quantized Variational Autoencoders (RQ-VAE) as a novel PEFT strategy within our framework. Extensive experiments across various FR model backbones and datasets demonstrate that our framework significantly reduces communication overhead while improving accuracy. 
The source code is available at \url{https://github.com/young1010/FedPEFT}.
\end{abstract}


\section{Introduction}
Due to growing concerns over user privacy in centralized recommendation systems and strict data protection regulations such as the 
EU General Data Protection Regulation~\cite{voigt2017eu}, Federated Recommendation (FR)~\cite{DBLP:series/lncs/YangTZCY20} has emerged as a distributed learning framework in which users' raw data remains inaccessible to the central server. The core idea of FR is that each user keeps their data on a local client, trains a local model, and only shares the model parameters with a central server, which then aggregates them and distributes the updated global model back to users.

However, to achieve high recommendation accuracy, most FR models~\cite{DBLP:journals/expert/ChaiWCY21, DBLP:conf/ijcai/ZhangL0YZZY23, DBLP:conf/aaai/AgrawalSKJ24, DBLP:conf/aaai/WangBHLWL25} rely mainly on item embeddings, whose size grows linearly with the number of items.
In many models, item embeddings constitute the majority of parameters, often exceeding those in the neural network components~\cite{yang2024hyperbolic, yang2024hgformer, xu2024aligngroup, yuan2025pkgrec, Darec, TTTRec}.
As a result, in scenarios with a large item set, the communication cost becomes a major bottleneck due to the massive embedding parameters.

\begin{figure}[!t]
  \centering
    \includegraphics[width=0.38\textwidth]{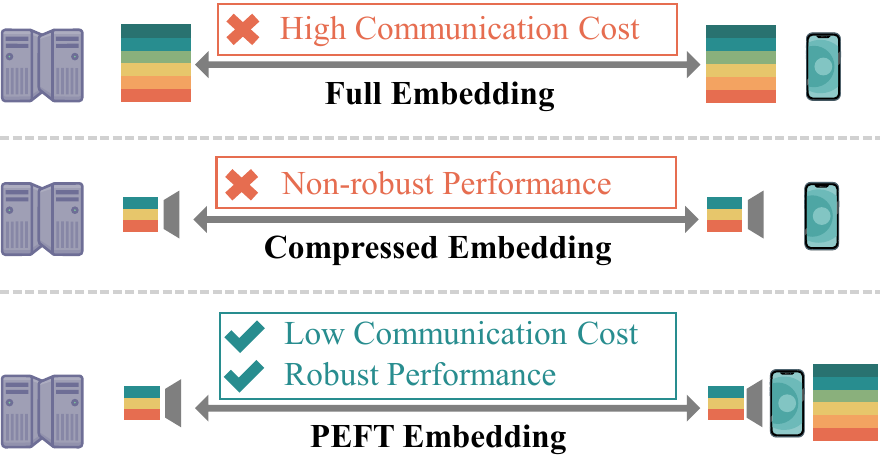}
  \caption{Comparison of full embeddings, compressed embeddings only, and PEFT based embeddings in FR.}
  \label{fig.1}
\end{figure}

To address this challenge, recent research has explored efficient embedding techniques, such as low-rank factorization~\cite{DBLP:conf/sigir/LinRCRY0RC20}, hash~\cite{DBLP:journals/tnn/ZhangZSL25}, and quantization~\cite{DBLP:journals/tois/ZhangLWHL23}, to compress the redundant parameters in full item embeddings. However, these methods often suffer from a noticeable drop in recommendation accuracy compared to models using full embeddings. To mitigate this performance gap, more complex architectures such as meta-learning~\cite{DBLP:conf/icml/FinnAL17} and Squeeze and Excitation Network (SENet)~\cite{DBLP:conf/cvpr/HuSS18} have been introduced. While these approaches can improve accuracy in some FR models, they also significantly reduce robustness when applied across diverse FR models and settings.
The above-mentioned limitations highlight the urgent need for a lightweight FR training framework that eliminates dependence on additional complex models.

To this end, our paper proposes a novel FR framework that leverages Parameter-Efficient Fine-Tuning (PEFT) for item embeddings. Our approach combines full embeddings with compressed embeddings to reduce communication overhead while enhancing recommendation performance, as illustrated in Figure~\ref{fig.1}. The proposed framework is model-agnostic and can be seamlessly plugged into existing embedding-based FR methods. In this framework, the server first pre-trains the full item embeddings using item attributes. We conduct a brief warm-up (<20 of 1000 rounds) training full embedding to stabilize early optimization. Afterwards, the full embeddings are frozen locally, and only the compressed embeddings are fine-tuned and transmitted until training is complete. 
This design enables the FR systems to achieve more efficient communication while maintaining robust and improved recommendation performance.

To further investigate the underlying effectiveness of our framework, we incorporate two existing compressed embedding strategies, aiming to identify the most suitable approach under different FR settings. Additionally, we explore the use of Residual Quantized Variational Autoencoders (RQ-VAE)~\cite{DBLP:conf/cvpr/LeeKKCH22, DBLP:conf/nips/RajputMSKVHHT0S23} as a novel strategy, which encodes each item into multiple codebooks with quantized semantic codes using residual vectors. In our framework, the semantic codes are pre-trained on the server and kept frozen during federated training, while the codebooks are fine-tuned. This design provides a compact yet expressive representation for item embeddings.

We evaluate the effectiveness of our method on various backbone models and public datasets. Experimental results show that while compressed embeddings typically suffer from 
poor 
robustness, our method 
successfully 
improves recommendation performance while reducing the communication overhead, thereby achieving a better trade-off between efficiency and performance. Furthermore, to evaluate the effectiveness of our framework with differential privacy, we conduct experiments under different settings~\cite{DBLP:conf/icalp/Dwork06}. The results indicate that our method consistently achieves stable improvements across various settings.
In summary, the key contributions of this work are as follows:
\begin{itemize}
\item We propose a lightweight, model-agnostic FR framework that incorporates PEFT for item embeddings, which significantly reduces the size of updated embedding parameters in the whole FR process while improving recommendation performance.
\item We present three compressed embedding strategies tailored to diverse FR settings and datasets within our framework, two of which consistently outperform the full embedding baseline.
\item We introduce RQ-VAE as a novel PEFT method that encodes item embeddings into compact codebooks and quantized semantic codes, achieving superior performance compared to other PEFT approaches within the framework in many cases.
\end{itemize}

\section{Problem Formulation}
In the Federated Recommendation (FR) setting, the central server maintains a set of $n$ items $\mathcal{I} = \{i\}$, each associated with item attributes. There are $m$ users/clients $\mathcal{U} = \{u\}$, and each client $u$ holds only its private data $D_u = \{(i, r_{ui}) \mid i \in \mathcal{I}\}$. $r_{ui} = 1$ indicates a positive interaction between user $u$ and item $i$ and $r_{ui} = 0$ indicates a negative one. 

In each round of FR training, the server randomly selects a subset of clients $U_t \subseteq \mathcal{U}$ based on a sampling ratio $\mathcal{S}$, and distributes the current global model $\Theta = \{E, W_g\}$ to each selected client $u \in U_t$. Here, $E = \{e_i\}_{i=1}^{n}$ denotes the item embeddings where $e_i \in \mathbb{R}^{k}$, and $W_g$ represents the global recommendation model (if applicable). Each client then locally optimizes its model or gradients and uploads the updates to the server. The server performs federated aggregation of the collected updates to refine the global model for the next round. After training finishes, clients download the final global model from the server for inference.

\section{Methodology}
\begin{figure*}[!t]
  \centering
    \includegraphics[width=0.82\textwidth]{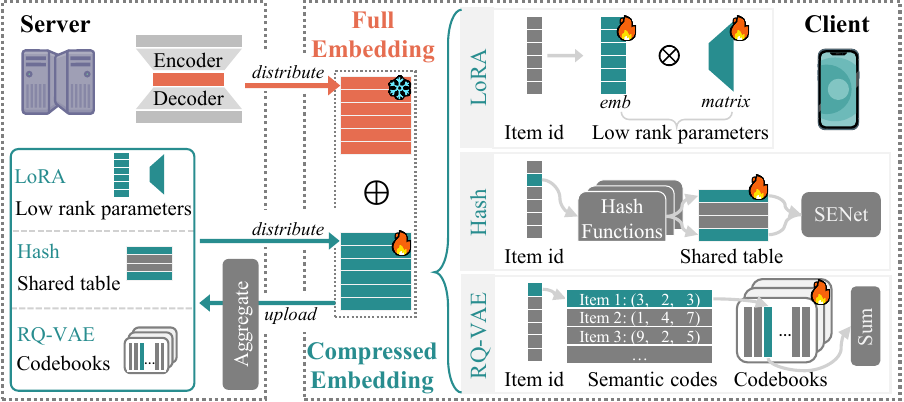}
  \caption{The plug-and-play FR framework that employs multiple compressed embedding strategies to implement PEFT.}
  \label{fig.3}
\end{figure*}

In this section, we propose an FR framework that incorporates Parameter-Efficient Fine-Tuning (PEFT) for item embeddings. Specifically, we first pre-train the full item embeddings using an AutoEncoder (AE) model. The pre-trained full embeddings are then distributed to clients and optimized for a few FR rounds. Afterwards, the server initializes and distributes compressed embeddings to clients. During subsequent training, clients freeze the full embeddings and only optimize and upload the compressed embeddings. The server only aggregates and distributes the compressed embeddings. The overall framework is illustrated in Figure~\ref{fig.3}.

\subsection{Pre-train}
Before pre-training the full item embeddings, we first encode the item attributes stored on the server into input embeddings $x \in \mathbb{R}^{k_p}$ using a publicly available pre-trained encoder. Then, we employ an unsupervised AE model, consisting of an encoder $\mathcal{E}_{\text{AE}}$ and a decoder $\mathcal{D}_{\text{AE}}$, to pre-train the item representations. Specifically, the input embedding is encoded into latent vector as $z_{\text{AE}} = \mathcal{E}_{\text{AE}}(x) \in \mathbb{R}^{k}$, and then reconstructed as $\hat{x}_{\text{AE}} = \mathcal{D}_{\text{AE}}(z_{\text{AE}})$. The objective is to minimize the reconstruction loss $\mathcal{L}_{\text{AE}} = \|x - \hat{x}_{AE}\|^2 $. After training the AE, the latent vector $z_{\text{AE}}$ is used as the initial full item embeddings $E = \{e_i \in \mathbb{R}^{k}\}_{i=1}^{n}$. 

Then, we optimize $E$ in FR training for only a few rounds as we observe that even limited updates can lead to noticeable performance improvements while having a negligible impact on the average communication overhead per round.

\subsection{Compressed Embedding}
After pre-training, the server distributes the initialized compressed embeddings to the clients. In each FR round, selected clients freeze the full item embedding $E$, and instead optimize and upload only the compressed embedding parameters for aggregation. 
We provide three compressed embedding strategies in our framework.

\vspace{0.4em}\noindent\textbf{Low-Rank Adaptation (LoRA)} is a widely used PEFT method that injects trainable low-rank decomposition matrices for fine-tuning~\cite{DBLP:conf/iclr/HuSWALWWC22, DBLP:conf/sigir/LinRCRY0RC20}. 

In our framework, we introduce a low-dimensional embedding table $A = \{\mathbf{a}_i \in \mathbb{R}^{k_{L}}\}_{i=1}^{n}$ and a low-rank matrix $B \in \mathbb{R}^{k \times k_{L}}$, where $k_{L} \ll k$. The compressed embedding vector $\mathbf{e}_i$ is then obtained by projecting $\mathbf{a}_i$ into the higher-dimensional space through matrix multiplication:
\begin{equation}\label{eq_lora}
   \mathbf{e}_i =  B(\mathbf{a}_i).
\end{equation}

We initialize the parameters where the embedding table $A$ follows initial distributions as the full embedding and the matrix $B = \mathbf{0}^{k \times k_{L}}$. 
In 
each FR training round, the server distributes $A$ and $B$ to the selected clients. The final item embedding in clients is computed as:
\begin{equation}\label{}
    \mathbf{E} = E + B(A) = \{e_i + B(\mathbf{a}_i)\}_{i=1}^{n}.
\end{equation}
Both the embedding table $A$ and the matrix $B$ are updated during local training on clients and uploaded to the server. 

\vspace{0.4em}\noindent\textbf{Hash} is a widely used compressed strategy that encodes each item into multiple vectors via hash functions, which are constructed to form the compressed item embedding~\cite{DBLP:conf/nips/SvenstrupHW17, DBLP:journals/pvldb/ZhangZMSLYC23}. 

In our framework, following~\cite{DBLP:journals/tnn/ZhangZSL25}, the server obtains a shared embedding table that is defined as $H = \{v_i\}_{i=1}^{d_{H}}$, where $d_{H} \ll n$, and a family of universal hash functions 
defined by $\mathcal{H}$. Each hash function $\mathcal{H}_{j}$ maps an item ID (integer inputs) into the index of a vector in $H$ as follows:
\begin{equation}
   \mathcal{H}_{j}(i) = \{[(a\cdot ID + b)\bmod p]\bmod d_{H}\} , 
\end{equation}
where $a$ and $b$ are randomly chosen integers with $a \ne 0$, $a \ne b$, and $p$ is a large prime number. A larger value of $p$ generally results in a lower probability of hash collisions. 

Before FR training, the server first randomly selects $h$ hash functions $\{\mathcal{H}_1, \dots, \mathcal{H}_h\}$ from the hash function family $\mathcal{H}$, and distributes them along with the initial shared embedding table $H = \{v_i\}_{i=1}^{d_{H}}$ to clients. Each item $i$ is presented by a set of $h$ vectors $\mathbf{v}_i = [\mathbf{v}_i^1, \dots , \mathbf{v}_i^h]$, where $\mathbf{v}_i^j = H[\mathcal{H}_{j}(i)] = v_{\mathcal{H}_{j}(i)}$ is the embedding vector at the hashed index form table $H$. Finally, the compressed embedding vector $\mathbf{e}_i$ can be constructed using one of the following methods: 

1) \textit{Mean}, only using the mean pooling $Mean(\cdot )$:
\begin{equation}\label{eq_hash_1}
   \mathbf{e}_i = Mean(\mathbf{v}_i^1, \dots , \mathbf{v}_i^h) = \frac{1}{h}\sum_{j=1}^{h}v_{\mathcal{H}_j(i)}.
\end{equation}

The final item embedding is:
\begin{equation}\label{}
    \mathbf{E} = \left\{e_i + \frac{1}{h}\sum_{j=1}^{h}v_{\mathcal{H}_j(i)} \right\}_{i=1}^{n}.
\end{equation}

2) \textit{Squeeze and Excitation Network (SENet)}, an attention mechanism designed to reweight multiple hash vectors dynamically with minimal additional parameters~\cite{DBLP:conf/cvpr/HuSS18}. 

The first step, known as \textit{squeeze}, aggregates the hash vectors into a $h$-dimension vector $\mathbf{s}_i = [s_i^1, \dots , s_i^h]$ with mean pooling as:
\begin{equation}\label{eq_hash_1}
   s_i^j = Mean(\mathbf{v}_i^j) = \frac{1}{k}\sum_{t=1}^k\mathbf{v}_i^{j,t} .
\end{equation}

The second step, referred to as the \textit{excitation}, learns the dynamic weights $\mathbf{w}_i = [w_i^1, \dots , w_i^h]$ using a two-layer multilayer perceptron (MLP) as:
\begin{equation}\label{eq_hash_2}
   \mathbf{w}_i = \sigma_2(\mathbf{W}_2(\sigma_1(\mathbf{W}_1\mathbf{s}_i)),
\end{equation}
where $\sigma_1$ is ReLU activation and $\sigma_2$ is the sigmoid activation. $\mathbf{W}_1 \in \mathbb{R}^{h_1 \times h}$ and $\mathbf{W}_2 \in \mathbb{R}^{h \times h_1}$ are two matrices in MLP, where the hidden dimension $h_1$ is typically determined by a expansion ratio $r_h = h_1/h$. In the following experiments, we set $r_h$ as $16$.

Then we obtain 
a weighted sum of hash vectors:
\begin{equation}\label{eq_hash_2}
   \mathbf{e}_i = \sum_{j=1}^h w_i^j \cdot \mathbf{v}_i^j = \sum_{j=1}^h w_i^j \cdot v_{\mathcal{H}_{j}(i)}.
\end{equation}

The final item embedding is:
\begin{equation}\label{}
    \mathbf{E} = \left\{e_i + \sum_{j=1}^h w_i^j \cdot v_{\mathcal{H}_{j}(i)} \right\}_{i=1}^{n}.
\end{equation}

In the FR training rounds, clients locally hold the frozen hash functions, and only optimize and upload the shared table. If SENet is used, the two matrices, $\mathbf{W}_1$ and $\mathbf{W}_2$, also need to be locally optimized and aggregated during training. However, these matrices are typically small and do not significantly increase the communication overhead.

\vspace{0.4em}\noindent\textbf{RQ-VAE}, called Residual Quantized Variational Autoencoders, is a multi-level vector quantization method applied to residuals. It has been primarily utilized in generative retrieval-based recommendation tasks~\cite{DBLP:conf/nips/RajputMSKVHHT0S23, DBLP:journals/corr/abs-2502-18965}. Leveraging its quantization mechanism, RQ-VAE decouples the embedding size from the number of items, offering a scalable and compact representation. This makes it a promising candidate 
as a compressed embedding method in our FR framework.

Specifically, there are $l$ shared embedding tables called codebooks $(C_0, \dots, C_{l-1} )$ and each item $i$ is represented by a semantic code, which is a tuple of indices $\mathbf{c_i} = (c_0, \dots , c_{l-1})$ of length $l$, where each index $c_j$ comes from the codebook $C_j = \{o_{j,i}\}_{i=1}^{d_{R}}$ and $d_{R} \ll n$. For example, if $c_j = i$, then $C_j(c_j) = o_{j,i} \in \mathbb{R}^{k}$.

In our framework, we first pre-train the semantic code in RQ-VAE before FR training on the server. Specifically, the item attributes (e.g., content features generally) are encoded into input embedding $x \in \mathbb{R}^{k_p}$ using a pre-trained encoder like sentence-t5~\cite{DBLP:conf/acl/NiACMHCY22}. Then RQ-VAE encodes the input $x$ by an encoder $\mathcal{E}$ to learn a latent embedding as $z = \mathcal{E}(x) \in \mathbb{R}^{k}$. At the zero-th level of RQ-VAE, the initial residual is $r_0 = z$. At each level $j$, finding the nearest vector $o_{j,t}$ to the current residual $r_j$ from the codebook $C_j$, and setting or updating the sematic code $c_j$ as the index of $o_{j,t}$ as $c_j = t = \arg\min||r_j- o_j||$. For the next level, the residual is $r_{j+1} = r_j - o_{j,{c_j}}$. After all levels are processed, the quantized representation of $z$ is obtained:
\begin{equation}\label{eq_rqvae}
    \hat{z} = \sum_{j=0}^{l-1}o_{j,{c_j}} = \sum_{j=0}^{l-1}C_j(c_j).
\end{equation} 
Then a decoder $\mathcal{D}$ reconstructs the input embedding $x$ via $\hat{x} = \mathcal{D}(\hat{z})$.

RQ-VAE jointly optimizes the $\mathcal{E}$, $\mathcal{D}$, and codebooks $(C_0, \dots, C_{l-1} )$ using the loss $\mathcal{L}(x) = \mathcal{L}_{\text{recon}} + \mathcal{L}_{\text{rqvae}}$, where $\mathcal{L}_{\text{recon}} = \|x - \hat{x}\|^2 $ and $\mathcal{L}_{\text{rqvae}} = \sum_{j=0}^{l-1} \left( \| \text{sg}[r_j] - o_{j,c_j} \|^2 + \beta \| r_j - \text{sg}[o_{j,c_j}] \|^2 \right)$.
Here, $\text{sg}[\cdot]$ denotes the stop-gradient operation, and $\beta$ controls the commitment loss weight. To reduce codebook collisions, RQ-VAE uses K-means clustering on the residuals $r_j$ at each level on the first training batch to initialize the codebooks with the corresponding centroids.

Then, the server distributes the pre-trained semantic codes to the clients. During the FR training phase, clients keep these semantic codes frozen. As for the codebooks, instead of using the pre-trained versions as initialization, we initialize them with the same distribution as the full embedding, consistent with the FR settings. The final compressed embedding is then obtained using the quantized representation from the compact codebooks $\mathbf{e}_i = \hat{z}$, defined as:
\begin{equation}\label{}
    \mathbf{E} = \left\{e_i + \sum_{j=0}^{l-1}C_j(c_j), \quad c_j \in \mathbf{c}_i \right\}_{i=1}^{n}.
\end{equation}

In each FR training round, only the codebooks are optimized in the clients and uploaded to the server.

\section{Experiments and Discussions}

\subsection{Experimental Setup}

\vspace{0.2em}
\noindent\textbf{Datasets.}
Three public recommendation datasets are used: MovieLens\mbox{-}1M~\cite{DBLP:journals/tiis/HarperK16}, which is denoted as ML1M, and two categories of the Amazon dataset~\cite{DBLP:conf/emnlp/NiLM19}, namely Software and Industrial. These datasets offer user\mbox{-}item interaction records along with item attributes for pre-training full embeddings. For instance, ML1M includes movie information, while the Amazon datasets provide item descriptions. 

\begin{table}[!t]
\centering
\resizebox{0.95\linewidth}{!}{
\begin{tabular}{lrrrr}
\toprule
\textbf{Dataset}   & \bf \# Users & \bf \# Items & \bf \# Interactions & \bf Sparsity \\
\midrule
ML1M        & 6,040            & 3,706            & 1,000,209                & 95.53\%           \\
Software & 1,826           & 802           & 12,805                 & 99.13\%           \\
Industrial     & 11,041          & 5,334           & 77,071                 & 99.87\%           \\
\bottomrule
\end{tabular}}
\caption{Dataset statistics.}
\label{tab.2}
\end{table}

\vspace{0.4em}\noindent\textbf{Backbones.}
We adopt four popular FR backbone models:
i) FedMF~\cite{DBLP:journals/expert/ChaiWCY21}, a federated matrix factorization with only user and item embeddings, 
ii) FedNCF~\cite{DBLP:journals/kbs/PerifanisE22}, the extension of FedMF with an MLP,
iii) FedPerGNN~\cite{wu2022federated}, a graph-based FR model using only user and item embeddings, and 
iv) PFedRec~\cite{DBLP:conf/ijcai/ZhangL0YZZY23}, a personalized FR model with item embeddings and client-specific MLPs
without user embeddings.






\vspace{0.4em}\noindent\textbf{Evaluation Metrics.} 
We adopt the widely used top-K recommendation evaluation metrics: Hit Ratio (H@K) and Normalized Discounted Cumulative Gain (N@K). Both metrics reflect the ranking quality, where higher values indicate better performance. 

\vspace{0.4em}\noindent\textbf{Implementation Details.}
In our FR setting, each client keeps its user embedding locally and does not upload it.
The client sampling ratio $S$ is set to $10\%$. Each selected client performs $2$ local epochs per round, and the total number of global rounds is $1,000$.

For the pre-training in the server, we use the open sentence-t5~\cite{DBLP:conf/acl/NiACMHCY22} to encode the item attributes into input embedding with dimension $k_p = 768$. The encoder $\mathcal{E}_{AE}$ consists of fully connected layers $[768, 512, 256, 128, 32]$ with ReLU activation and the decoder is reverse. Pre-training is conducted for a number of rounds chosen from $\{10^3, 10^4, 10^5, 10^6\}$, with a learning rate selected from $\{1e-3, 1e-4\}$.

For the compressed embeddings, we set the size of latent embedding $k_{L}$ in $\{2,3,4,5,6\}$ in LoRA.
In RQ-VAE, we set the loss weight $\beta =0.25$. The length of semantic codes $l$ is selected from $\{2,3,4,5,6\}$, The size of codebook $d_{R}$ is set in $\{32, 64, 128, 256, 512\}$.
In hash, the size of the shared embedding table $d_{H}$ is selected from $\{256, 512, 1024\}$, and the collision parameter $p$ is set as $4,096$. The number of hash functions $h$ is set in $\{1,2,3,4\}$.

Our implementation is based on the open\mbox{-}source library for recommendation, FuxiCTR~\cite{DBLP:conf/cikm/ZhuLYZH21}. All experiments are implemented using PyTorch and executed on machines equipped with NVIDIA GeForce RTX 3090 GPUs.

\begin{table*}[!t]
\setlength{\belowcaptionskip}{-4mm}
\resizebox{\linewidth}{!}{
\begin{tabular}{l|l|cccc|cccc|cccc}
\toprule
\multicolumn{1}{c|}{\multirow{2}{*}{\bf Method}}    & \multicolumn{1}{c|}{\multirow{2}{*}{\bf Embedding}} & \multicolumn{4}{c|}{\bf ML1M}                                          & \multicolumn{4}{c|} {\bf Software}                                      & \multicolumn{4}{c}{\bf Industrial}                                    \\ \cmidrule(lr){3-6} \cmidrule(lr){7-10} \cmidrule(lr){11-14} 
                           & \multicolumn{1}{c|}{}                           & N@10           & H@10           & N@20           & H@20           & N@10           & H@10           & N@20           & H@20           & N@10           & H@10           & N@20           & H@20           \\ \midrule
\multirow{9}{*}{FedMF}     & Full                                           & 33.98          & 58.44          & 38.27          & 75.98          & 21.42          & 29.75          & 24.16          & 40.65          & 9.13           & 16.48          & 11.94          & 27.70          \\ \cmidrule(lr){2-14}
                           & C-LoRA                                         & 36.66          & 58.08          & 41.01          & 75.22          & 16.20          & 22.23          & 18.51          & 31.43          & 8.00           & 14.10          & 10.57          & 24.38          \\
                           & C-RQ-VAE                                        & 8.62           & 16.56          & 11.96          & 29.85          & 15.51          & 22.73          & 18.05          & 32.86          & 9.82           & 18.08          & 12.68          & 27.69          \\
                           & C-Hash                                         & 5.54           & 12.19          & 8.30           & 23.26          & 11.50          & 19.17          & 14.00          & 29.24          & 5.25           & 11.09          & 7.75           & 21.07          \\
                           & C-Hash(S)                                      & 5.48           & 12.14          & 8.36           & 23.64          & 12.63          & 21.14          & 14.96          & 30.39          & 5.01           & 10.53          & 7.40           & 20.10          \\ \cmidrule(lr){2-14}
                           & P-LoRA                                         & \textbf{37.98} & \textbf{59.79} & \textbf{42.42} & \textbf{76.60} & 20.54          & 29.88          & 23.43          & 41.43          & 9.86           & 17.32          & 12.47          & 27.74          \\
                           & P-RQ-VAE                                        & 33.59          & 58.96          & 38.06          & 76.52          & \textbf{21.46} & \textbf{29.88} & \textbf{24.39} & \textbf{41.52} & \textbf{10.82} & \textbf{18.10} & \textbf{13.42} & \textbf{28.49} \\
                           & P-Hash                                         & 19.85          & 38.92          & 23.69          & 54.11          & 16.20          & 21.52          & 18.62          & 31.27          & 5.69           & 11.51          & 8.24           & 21.71          \\
                           & P-Hash(S)                                      & 16.69          & 33.79          & 20.37          & 48.36          & 15.66          & 20.97          & 17.84          & 29.68          & 5.28           & 10.98          & 7.78           & 20.95          \\ \midrule
\multirow{9}{*}{FedNCF}    & Full                                           & 38.80          & \textbf{61.29} & 43.08          & 78.11          & 17.07          & \textbf{24.27} & 19.38          & 33.54          & 8.99           & 15.62          & 11.47          & 25.53          \\ \cmidrule(lr){2-14}
                           & C-LoRA                                         & 35.03          & 51.94          & 40.73          & 74.59          & 16.79          & 23.06          & 19.25          & 32.86          & 6.47           & 12.67          & 8.89           & 22.35          \\
                           & C-RQ-VAE                                        & 14.76          & 25.40          & 18.51          & 40.51          & 14.77          & 20.92          & 17.19          & 30.50          & 7.45           & 12.90          & 9.92           & 22.78          \\
                           & C-Hash                                         & 6.44           & 13.28          & 8.88           & 23.13          & 12.88          & 20.97          & 15.08          & 29.79          & 6.20           & 11.77          & 8.71           & 21.80          \\
                           & C-Hash(S)                                      & 8.52           & 23.92          & 12.85          & 40.78          & 12.07          & 19.55          & 14.08          & 27.60          & 6.87           & 13.45          & 9.35           & 23.39          \\ \cmidrule(lr){2-14}
                           & P-LoRA                                         & 39.52          & 60.72          & 43.97          & 78.28          & 17.08          & 24.09          & 19.38          & 33.21          & 9.61           & 16.99          & 12.07          & 26.82          \\
                           & P-RQ-VAE                                        & \textbf{39.75} & 60.91          & \textbf{44.13} & \textbf{78.28} & \textbf{17.09} & 24.06          & \textbf{19.53} & \textbf{33.81} & 9.51           & 16.75          & 12.05          & 26.89          \\
                           & P-Hash                                         & 36.89          & 58.43          & 41.57          & 76.66          & 17.00          & 23.99          & 18.86          & 31.33          & 9.60           & 16.93          & 12.21          & 27.35          \\
                           & P-Hash(S)                                      & 37.94          & 58.99          & 42.52          & 76.97          & 17.09          & 23.93          & 19.00          & 31.54          & \textbf{9.88}  & \textbf{17.30} & \textbf{12.45} & \textbf{27.58} \\ \midrule
\multirow{9}{*}{FedPerGNN} & Full                                           & 27.24          & 50.43          & 33.72          & 75.95          & 21.89          & 30.54          & 24.60          & 41.33          & 8.65           & 14.81          & 11.21          & 25.04          \\ \cmidrule(lr){2-14}
                           & C-LoRA                                         & \textbf{30.73} & \textbf{57.04} & \textbf{35.78} & 76.85          & 16.32          & 22.29          & 19.05          & 33.19          & 7.39           & 13.53          & 9.85           & 23.36          \\
                           & C-RQ-VAE                                        & 11.94          & 26.36          & 16.67          & 45.08          & \textbf{25.23} & \textbf{38.12} & \textbf{29.02} & \textbf{53.07} & \textbf{13.15} & \textbf{24.54} & \textbf{16.04} & \textbf{35.99} \\
                           & C-Hash                                         & 7.30           & 16.26          & 12.72          & 37.98          & 15.13          & 22.29          & 17.46          & 31.54          & 6.63           & 12.81          & 9.10           & 22.70          \\
                           & C-Hash(S)                                      & 7.99           & 18.01          & 12.94          & 37.68          & 15.45          & 22.95          & 17.88          & 32.64          & 6.76           & 13.11          & 9.29           & 23.16          \\ \cmidrule(lr){2-14}
                           & P-LoRA                                         & 29.36          & 55.15          & 34.62          & 75.81          & 18.84          & 27.06          & 21.89          & 39.22          & 7.36           & 13.51          & 9.93           & 23.81          \\
                           & P-RQ-VAE                                        & 30.10          & 56.35          & 35.58          & \textbf{77.84} & 23.45          & 36.52          & 27.22          & 51.52          & 12.08          & 22.08          & 15.11          & 34.12          \\
                           & P-Hash                                         & 29.18          & 55.56          & 34.50          & 76.56          & 22.38          & 33.35          & 25.74          & 46.66          & 9.13           & 15.54          & 11.67          & 25.69          \\
                           & P-Hash(S)                                      & 29.67          & 56.26          & 34.92          & 76.97          & 21.94          & 31.43          & 24.82          & 42.77          & 8.42           & 14.77          & 10.93          & 24.82          \\ \midrule
\multirow{9}{*}{PFedRec}   & Full                                           & 38.63          & 60.48          & 42.93          & 77.84          & 17.10          & 24.25          & 19.42          & 33.49          & 8.90           & 15.56          & 11.39          & 25.50          \\ \cmidrule(lr){2-14}
                           & C-LoRA                                         & 28.02          & 48.96          & 33.51          & 71.13          & 16.66          & 22.89          & 19.28          & 33.35          & 7.84           & 13.89          & 10.39          & 24.14          \\
                           & C-RQ-VAE                                        & 12.88          & 28.48          & 17.70          & 47.73          & 16.72          & 24.70          & 19.15          & 34.34          & 7.86           & 13.97          & 10.45          & 24.35          \\
                           & C-Hash                                         & 5.59           & 13.96          & 9.25           & 28.54          & 13.11          & 20.59          & 16.05          & 32.42          & 6.32           & 12.45          & 8.66           & 21.79          \\
                           & C-Hash(S)                                      & 6.19           & 11.21          & 8.75           & 21.47          & 15.35          & 21.30          & 17.73          & 30.78          & 6.64           & 12.47          & 9.19           & 22.69          \\ \cmidrule(lr){2-14}
                           & P-LoRA                                         & \textbf{39.48} & \textbf{61.35} & \textbf{43.88} & \textbf{78.25} & \textbf{17.28} & \textbf{24.49} & \textbf{19.64} & \textbf{33.91} & 9.62           & 16.97          & 12.07          & 26.77          \\
                           & P-RQ-VAE                                        & 38.89          & 59.40          & 43.55          & 77.84          & 17.12          & 24.05          & 19.51          & 33.66          & 9.32           & 16.49          & 11.71          & 26.06          \\
                           & P-Hash                                         & 37.92          & 59.11          & 42.19          & 75.81          & 17.11          & 24.04          & 19.14          & 32.15          & 9.62           & 17.05          & 12.29          & 27.64          \\
                           & P-Hash(S)                                      & 37.09          & 56.90          & 42.26          & 77.17          & 17.22          & 24.32          & 19.07          & 31.71          & \textbf{9.74}  & \textbf{17.17} & \textbf{12.38} & \textbf{27.71} \\ \bottomrule
\end{tabular}}
\caption{Recommendation performance of full embeddings (Full), compressed embeddings only (C), and PEFT embeddings (P) using various strategies including LoRA, RQ-VAE, Hash (with Mean), and Hash(S) (with SENet).} 
\label{tab_results}
\end{table*}

\subsection{Experimental Results}
Based on the averaged results from five runs, as shown in Table~\ref{tab_results}, we can make the following observations.

\vspace{0.4em}\noindent\textbf{PEFT embeddings consistently outperform in most scenarios.}
While compressed embeddings occasionally perform well, e.g., achieving the best results on the FedPerGNN, they also suffer from significant performance degradation in other settings, such as in the PFedRec-ML1M setting. Among them, C-LoRA and C-RQ-VAE generally outperform the C-Hash variants but still fall short of full embeddings in some settings. In contrast, PEFT embeddings demonstrate strong and consistent performance across FedMF, FedNCF, and PFedRec. Even on FedPerGNN, P-RQ-VAE also shows better performance than full embedding. 
Notably, on FedMF, P-LoRA and P-RQ-VAE show substantial improvements over full embeddings.

\vspace{0.4em}\noindent\textbf{Strategy effectiveness varies across models and datasets.}
Within the PEFT embeddings, P-RQ-VAE delivers stable and competitive results, sometimes even outperforming full embeddings. For example, in the FedMF-Industrial setting, it achieves the best overall performance. 
P-LoRA also 
shows 
robust performance results across multiple settings. Although it underperforms in FedPerGNN-Software and FedPerGNN-Industrial, 
P-LoRA
significantly outperforms others in FedMF-ML1M. 
The P-Hash strategies perform the best in FedNCF-Industrial and PFedRec-Industrial, particularly when combined with SENet. However, they struggle on FedMF, where performance drops notably. 
For compressed embeddings only, C-Hash exhibits poor and unstable performance in most cases. A likely reason is that it relies on random hashing. In contrast, C-LoRA and C-RQ-VAE generally outperform C-Hash, though they still lag behind full embeddings, with exceptions in FedMF and FedPerGNN. 

\vspace{0.4em}\noindent\textbf{SENet improves performance when combined with MLP.}
Focusing on PEFT embeddings with hash, we observe that SENet brings performance gains in models with the MLP architecture, such as FedNCF and PFedRec. In these settings, P-Hash(S) outperforms P-Hash. However, in models that rely solely on embeddings such as FedMF and FedPerGNN, SENet actually leads to a performance drop. For compressed embeddings only, SENet improves the performance in most settings.

\subsection{Communication Analysis}
We conduct a comprehensive analysis of the strategies in terms of client-side communication overhead on embedding, additional storage, additional computation cost, and embedding representation capacity. The results with the number of items $n$, embedding dimension $k$ are summarized in Table~\ref{tab_ananlysis}. 

From the table, we can observe that LoRA effectively avoids representation collisions due to its direct use of learnable embeddings, but incurs linear communication overhead with respect to the number of items $n$. RQ-VAE achieves a compact communication footprint and storage efficiency, while offering a large representation space via multi-level codebooks. However, its resistance to representation collision depends heavily on the quality of pre-training conducted on the server. In contrast, hash-based methods inherently mitigate collisions through the design of hash functions, but require a larger hash table size $d_{H}$ compared to the codebook size $d_{R} \cdot l$ used in RQ-VAE to achieve comparable representation capacity.

\begin{table}[!t]
\resizebox{\linewidth}{!}{
\begin{tabular}{l|cccc}
\toprule
 & \bf Communication & \bf Storage & \bf Computation & \bf Representation \\ \midrule
Full & $O(k \cdot n)$ & $O(k \cdot n)$ & $0$ & $n$\\ 
LoRA & $O(k_{L} \cdot (n+k))$ & $O(k_{L} \cdot (n+k))$ & $0$ & $n$\\ 
RQ-VAE & $O(d_{R} \cdot l)$ & $O(d_{R} \cdot l + n)$ & $0$ & $(d_{R})^l$\\
Hash & $O(d_{H})$ & $O(d_{H}+h)$ & $0$ & $C_{d_{H}+h-1}^h$\\
Hash(S) & $O(d_{H})$ & $O(d_{H}+h)$ & $O(h^2)$ & $C_{d_{H}+h-1}^h$\\
\bottomrule
\end{tabular}}
\caption{Analysis on embedding.}\label{tab_ananlysis}
\end{table}

\begin{figure}[!t]
  \centering
    \includegraphics[width=0.40\textwidth]{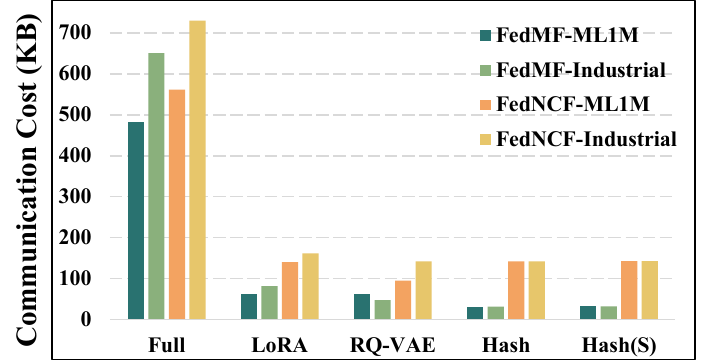}
  \caption{Average communication cost per client.}
  \label{fig.7}
\end{figure}

As 
shown in Figure~\ref{fig.7}, all PEFT embedding strategies substantially reduce communication overhead compared to full embeddings. Among them, hash-based methods achieve the greatest reduction overall, but do not show a significant reduction compared to other strategies. 
However, when considering recommendation performance, LoRA and RQ-VAE generally provide a more favorable trade-off in most cases, achieving a well-balanced compromise between effectiveness and communication efficiency.

\subsection{Hyperparameters Sensitivity}
To explore the trade-off between recommendation performance and communication in PEFT-based strategies, we analyze the impact of key hyperparameters for each method.

\vspace{0.4em}
\noindent\textbf{LoRA.} The primary hyperparameter is the latent dimension size $k_{L}$ in low-rank matrices. As shown in Table~\ref{tab_lora}, we evaluate PFedRec on the ML1M and Industrial datasets under varying values of $k_{L}$. Although the number of trainable parameters and communication overhead increase with larger $k_{L}$, the recommendation performance does not consistently improve. On ML1M, both N@10 and H@10 achieve their highest scores when $k_{L}=4$, even surpassing the performance of the full embedding baseline. However, when $k_{L}=5$ or $6$, performance degrades and falls below that of the full embedding. On the Industrial dataset, although all configurations of $k_{L}$ outperform the full embedding baseline, the best results 
are also obtained at $k_{L}=4$.

\begin{table}[!t]
\centering
\resizebox{\linewidth}{!}{
\begin{tabular}{c|ccc|ccc}
\toprule
\multirow{2}{*}{\bf $k_{L}$} & \multicolumn{3}{c|}{\bf ML1M} & \multicolumn{3}{c}{\bf Industrial} \\ \cmidrule(lr){2-4} \cmidrule(lr){5-7}
 & N@10 & H@10 & Comm.(KB) & N@10 & H@10 & Comm.(KB) \\
\midrule
2 & 38.16 & 59.19 & 30.1  & 9.58 & 16.91 & 40.7 \\
3 & 37.96 & 58.50 & 45.2  & 9.56 & 16.88 & 61.0 \\
4 & \textbf{39.48} & \textbf{61.35} & 60.3  & \textbf{9.62} & \textbf{16.97} & 81.4 \\
5 & 37.99 & 58.51 & 75.4  & 9.54 & 16.85 & 101.7 \\
6 & 37.88 & 58.19 & 90.5  & 9.56 & 16.90 & 122.1 \\
\midrule
Full & 38.63 & 60.48 & 482.4 & 8.90 & 15.56 & 651.1 \\
\bottomrule
\end{tabular}}
\caption{Performance and efficiency (Comm.) comparison on PFedRec under various $k_{L}$ values in LoRA strategy.}\label{tab_lora}
\end{table}

\begin{figure}[!t]
  \centering
    \subfigure[FedNCF-ML1M]{\includegraphics[width=0.45\textwidth]{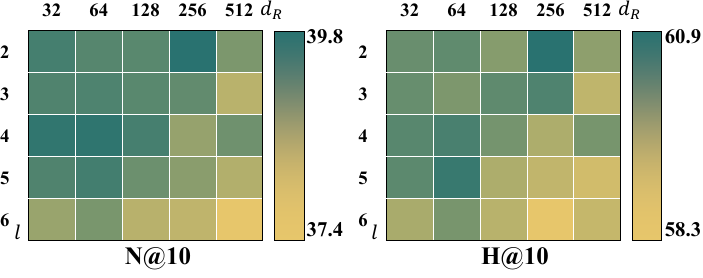}\label{fig.4.1}}
    \subfigure[FedPerGNN-Industrial]{\includegraphics[width=0.45\textwidth]{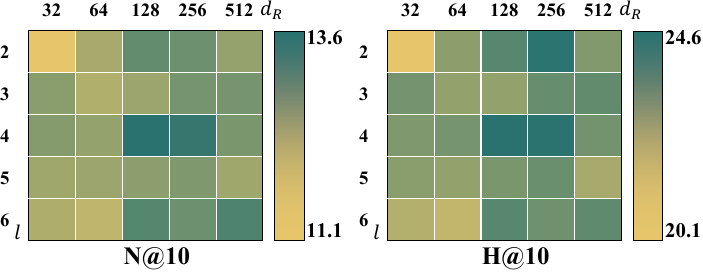}\label{fig.4.2}}
  \caption{Sensitivity analysis on RQ-VAE.}
  \label{fig.4}
\end{figure}

\vspace{0.4em}
\noindent\textbf{RQ-VAE.} RQ-VAE is controlled by two critical hyperparameters: the codebook size $d_{R}$ and the number of codebooks $l$. As shown in Figure~\ref{fig.4}, we evaluate the PR-VAE strategy on FedNCF-ML1M and FedPerGNN-Industrial settings under varying values of $d_{R}$ and $l$. We observe that increasing these values does not always lead to better performance. For example, when $d_{R}=512$, the overall performance is lower than that when $d_{R}=256$. A similar trend is observed with $l$. The performance begins to decline when $l=6$ in Figure~\ref{fig.4}, 
suggesting that overly large codebooks or too many quantization levels may introduce redundancy, which can weaken the relevance between items with similar semantic information and degrade representation quality.

\begin{figure}[!t]
  \centering
    \subfigure[N@10 FedNCF-ML1M]{\includegraphics[width=0.45\textwidth]{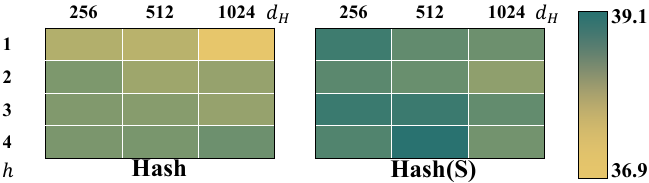}\label{fig.5.1}}
    \subfigure[N@10 FedNCF-Industrial]{\includegraphics[width=0.45\textwidth]{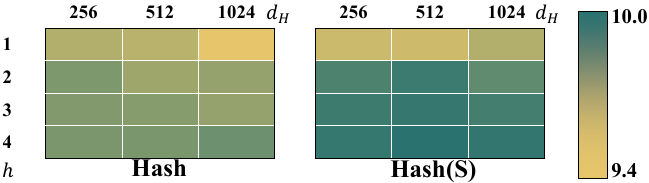}\label{fig.5.2}}
  \caption{Sensitivity analysis on Hash strategies.}
  \label{fig.5}
\end{figure}

\vspace{0.4em}
\noindent\textbf{Hash.} For the hash-based strategies, the main hyperparameters are the size of the shared embedding table $d_{H}$ and the number of hash functions $h$. As shown in Figure~\ref{fig.5}, we evaluate FedNCF on 
the ML1M and Industrial datasets under varying values of $d_{H}$ and $h$. The results show that Hash(S) (with SENet) consistently outperforms Hash (with Mean) when using FedNCF as the backbone model. Furthermore, performance improves with increasing $h$, especially from $h = 1$ to $h = 2$. However, when $h$ increases to 3 or 4, the performance gain becomes marginal. In contrast, increasing $d_{H}$ does not consistently lead to better performance. In many cases, smaller values of $d_{H}$ achieve competitive performance while maintaining lower communication cost.

\subsection{Analysis of DP Levels}
\begin{figure}[!t]
  \centering
    \subfigure[Industrial N@10]{\includegraphics[width=0.23\textwidth]{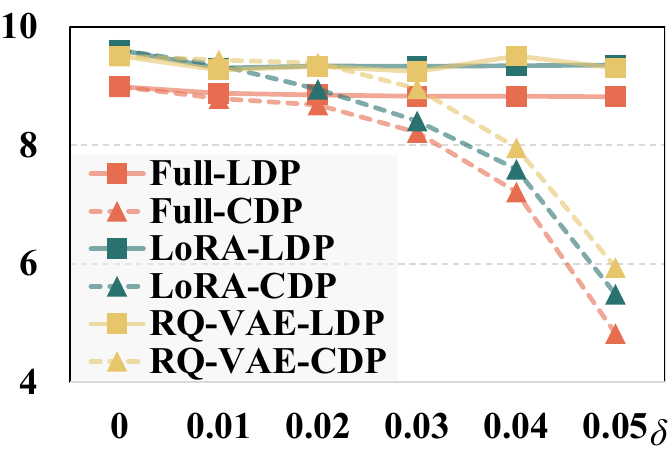}\label{fig.6.1}}
    \subfigure[Industrial H@10]{\includegraphics[width=0.23\textwidth]{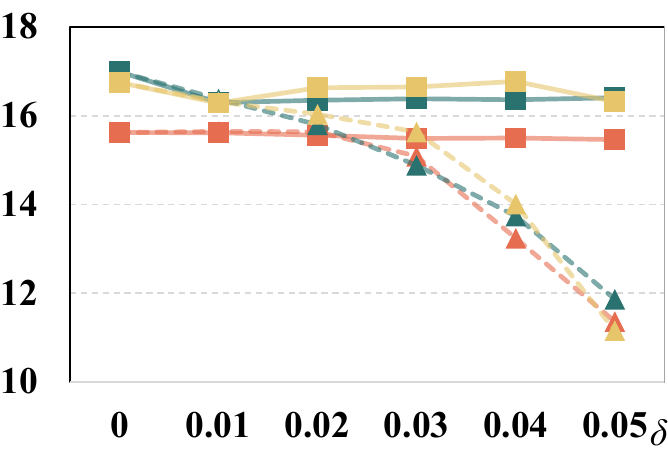}\label{fig.6.2}}
    \subfigure[ML1M  N@10]{\includegraphics[width=0.23\textwidth]{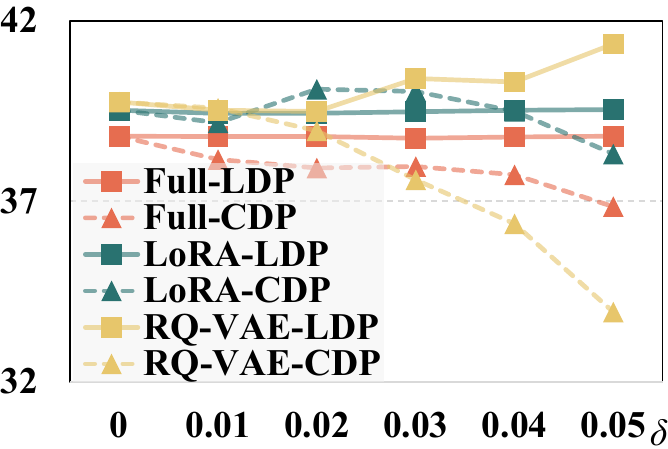}\label{fig.6.3}}
    \subfigure[ML1M  H@10]{\includegraphics[width=0.23\textwidth]{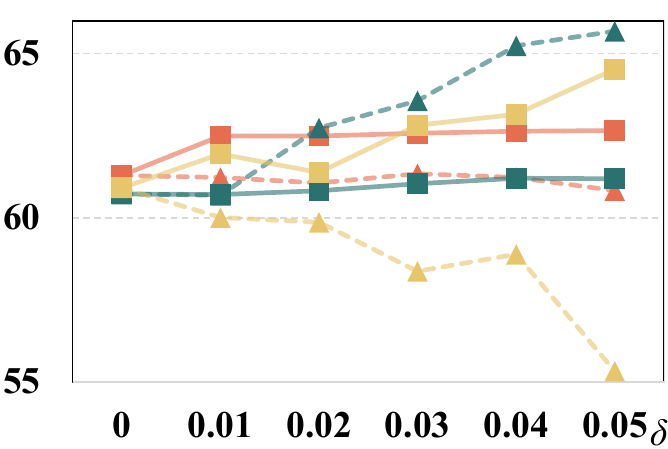}\label{fig.6.4}}
  \caption{Performance under different scale $\delta$ with CDP and LDP using the FedNCF model.}
  \label{fig.6}
\end{figure}

To assess the performance with differential privacy (DP), we conduct experiments under two widely studied DP settings~\cite{DBLP:conf/ndss/NaseriHC22}: Central Differential Privacy (CDP)~\cite{DBLP:journals/tifs/WeiLDMYFJQP20} and Local Differential Privacy (LDP)~\cite{DBLP:journals/iotj/ZhaoBSY25}. In both settings, we adopt the Laplace mechanism as the privacy-preserving strategy with the parameter $\delta$ that denotes the scale of the Laplace distribution. A larger $\delta$ introduces more noise, thus providing stronger privacy guarantees. The results are in Figure~\ref{fig.6}.

From the results on the Industrial dataset, both full embeddings and PEFT-based embeddings exhibit minimal performance degradation as $\delta$ increases under LDP. Under CDP, however, performance 
declines with increasing $\delta$. Notably, PEFT-based methods, including LoRA and RQ-VAE, consistently outperform full embeddings, indicating higher robustness in privacy-constrained scenarios. Overall, PEFT embeddings demonstrate superior performance and robustness compared to full embeddings on the Industrial dataset.

For the ML1M dataset, a similar trend is observed under LDP, with RQ-VAE even showing performance improvement as $\delta$ increases. However, under CDP, RQ-VAE suffers significant performance degradation, while the performance of full embeddings remains relatively stable. Interestingly, LoRA shows increased H@10 values under higher $\delta$, suggesting that LoRA is more robust under CDP, while RQ-VAE exhibits 
better 
performance under LDP.

\section{Conclusion}
This paper proposes a communication-efficient federated recommendation (FR) framework that leverages plugin Parameter-Efficient Fine-Tuning (PEFT) strategies for item embeddings. By freezing the pre-trained full item embeddings on the client side and only updating lightweight compressed embeddings, our method significantly reduces communication overhead while maintaining or even improving recommendation performance. 
Our framework adopts multiple compressed embedding strategies, including LoRA, Hash, and RQ-VAE, with these strategies applied to PEFT-based recommendation for the first time to the best of our knowledge.
Extensive evaluations across various FR models and datasets validate the effectiveness of our approach.

While our approach demonstrates strong empirical performance and adaptability, including under differential privacy settings, it also has certain limitations. No single strategy consistently outperforms others across all datasets and FR models. Additionally, although PEFT methods focus on reducing communication costs, compressed embedding techniques offer complementary advantages in terms of lower local storage requirements. These observations 
uncover a promising future direction in exploring the trade-offs between performance, communication efficiency, and client storage cost, potentially through hybrid approaches
.
\appendix

\section{Acknowledgments}
This work was supported by the National Natural Science Foundation of China under Grant 62372140.  
This funding applies only to Mr. Haochen Yuan, Dr. Xiang He and Prof. Zhongjie Wang, who are affiliated with Harbin Institute of Technology.  
Dr. Yang Zhang from the University of North Texas and Prof. Quan Z. Sheng from Macquarie University did not receive financial support for this work from this grant or from any other external project; their contributions were conducted as independent academic research.  
The authors sincerely thank all colleagues from the participating universities for their valuable collaboration and insights.
\bigskip

\bibliography{aaai2026}

\begin{thebibliography}{33}
\providecommand{\natexlab}[1]{#1}

\bibitem[{Agrawal et~al.(2024)Agrawal, Sirohi, Kumar, and Jayadeva}]{DBLP:conf/aaai/AgrawalSKJ24}
Agrawal, N.; Sirohi, A.~K.; Kumar, S.; and Jayadeva. 2024.
\newblock No Prejudice! Fair Federated Graph Neural Networks for Personalized Recommendation.
\newblock In \emph{Thirty-Eighth {AAAI} Conference on Artificial Intelligence, {AAAI} 2024}, 10775--10783.

\bibitem[{Chai et~al.(2021)Chai, Wang, Chen, and Yang}]{DBLP:journals/expert/ChaiWCY21}
Chai, D.; Wang, L.; Chen, K.; and Yang, Q. 2021.
\newblock Secure Federated Matrix Factorization.
\newblock \emph{{IEEE} Intell. Syst.}, 36: 11--20.

\bibitem[{Deng et~al.(2025)Deng, Wang, Cai, Ren, Hu, Ding, Luo, and Zhou}]{DBLP:journals/corr/abs-2502-18965}
Deng, J.; Wang, S.; Cai, K.; Ren, L.; Hu, Q.; Ding, W.; Luo, Q.; and Zhou, G. 2025.
\newblock OneRec: Unifying Retrieve and Rank with Generative Recommender and Iterative Preference Alignment.
\newblock arXiv:2502.18965.

\bibitem[{Dwork(2006)}]{DBLP:conf/icalp/Dwork06}
Dwork, C. 2006.
\newblock Differential Privacy.
\newblock In \emph{Automata, Languages and Programming, 33rd International Colloquium, {ICALP} 2006}, volume 4052, 1--12.

\bibitem[{Finn, Abbeel, and Levine(2017)}]{DBLP:conf/icml/FinnAL17}
Finn, C.; Abbeel, P.; and Levine, S. 2017.
\newblock Model-Agnostic Meta-Learning for Fast Adaptation of Deep Networks.
\newblock In \emph{Proceedings of the 34th International Conference on Machine Learning, {ICML} 2017}, volume~70, 1126--1135.

\bibitem[{Harper and Konstan(2016)}]{DBLP:journals/tiis/HarperK16}
Harper, F.~M.; and Konstan, J.~A. 2016.
\newblock The MovieLens Datasets: History and Context.
\newblock \emph{{ACM} Trans. Interact. Intell. Syst.}, 5: 19:1--19:19.

\bibitem[{Hu et~al.(2022)Hu, Shen, Wallis, Allen{-}Zhu, Li, Wang, Wang, and Chen}]{DBLP:conf/iclr/HuSWALWWC22}
Hu, E.~J.; Shen, Y.; Wallis, P.; Allen{-}Zhu, Z.; Li, Y.; Wang, S.; Wang, L.; and Chen, W. 2022.
\newblock LoRA: Low-Rank Adaptation of Large Language Models.
\newblock In \emph{The Tenth International Conference on Learning Representations, {ICLR} 2022}.

\bibitem[{Hu, Shen, and Sun(2018)}]{DBLP:conf/cvpr/HuSS18}
Hu, J.; Shen, L.; and Sun, G. 2018.
\newblock Squeeze-and-Excitation Networks.
\newblock In \emph{2018 {IEEE} Conference on Computer Vision and Pattern Recognition, {CVPR} 2018}, 7132--7141.

\bibitem[{Lee et~al.(2022)Lee, Kim, Kim, Cho, and Han}]{DBLP:conf/cvpr/LeeKKCH22}
Lee, D.; Kim, C.; Kim, S.; Cho, M.; and Han, W. 2022.
\newblock Autoregressive Image Generation using Residual Quantization.
\newblock In \emph{{IEEE/CVF} Conference on Computer Vision and Pattern Recognition, {CVPR} 2022}, 11513--11522.

\bibitem[{Lin et~al.(2020)Lin, Ren, Chen, Ren, Yu, Ma, de~Rijke, and Cheng}]{DBLP:conf/sigir/LinRCRY0RC20}
Lin, Y.; Ren, P.; Chen, Z.; Ren, Z.; Yu, D.; Ma, J.; de~Rijke, M.; and Cheng, X. 2020.
\newblock Meta Matrix Factorization for Federated Rating Predictions.
\newblock In \emph{Proceedings of the 43rd International {ACM} {SIGIR} conference on research and development in Information Retrieval, {SIGIR} 2020}, 981--990.

\bibitem[{Naseri, Hayes, and Cristofaro(2022)}]{DBLP:conf/ndss/NaseriHC22}
Naseri, M.; Hayes, J.; and Cristofaro, E.~D. 2022.
\newblock Local and Central Differential Privacy for Robustness and Privacy in Federated Learning.
\newblock In \emph{29th Annual Network and Distributed System Security Symposium, {NDSS} 2022}.

\bibitem[{Ni et~al.(2022)Ni, {\'{A}}brego, Constant, Ma, Hall, Cer, and Yang}]{DBLP:conf/acl/NiACMHCY22}
Ni, J.; {\'{A}}brego, G.~H.; Constant, N.; Ma, J.; Hall, K.~B.; Cer, D.; and Yang, Y. 2022.
\newblock Sentence-T5: Scalable Sentence Encoders from Pre-trained Text-to-Text Models.
\newblock In \emph{Findings of the Association for Computational Linguistics: {ACL} 2022}, 1864--1874.

\bibitem[{Ni, Li, and McAuley(2019)}]{DBLP:conf/emnlp/NiLM19}
Ni, J.; Li, J.; and McAuley, J.~J. 2019.
\newblock Justifying Recommendations using Distantly-Labeled Reviews and Fine-Grained Aspects.
\newblock In \emph{Proceedings of the 2019 Conference on Empirical Methods in Natural Language Processing and the 9th International Joint Conference on Natural Language Processing, {EMNLP-IJCNLP} 2019}, 188--197.

\bibitem[{Perifanis and Efraimidis(2022)}]{DBLP:journals/kbs/PerifanisE22}
Perifanis, V.; and Efraimidis, P.~S. 2022.
\newblock Federated Neural Collaborative Filtering.
\newblock \emph{Knowl. Based Syst.}, 242: 108441.

\bibitem[{Rajput et~al.(2023)Rajput, Mehta, Singh, Keshavan, Vu, Heldt, Hong, Tay, Tran, Samost, Kula, Chi, and Sathiamoorthy}]{DBLP:conf/nips/RajputMSKVHHT0S23}
Rajput, S.; Mehta, N.; Singh, A.; Keshavan, R.~H.; Vu, T.; Heldt, L.; Hong, L.; Tay, Y.; Tran, V.~Q.; Samost, J.; Kula, M.; Chi, E.~H.; and Sathiamoorthy, M. 2023.
\newblock Recommender Systems with Generative Retrieval.
\newblock In \emph{Advances in Neural Information Processing Systems 36: Annual Conference on Neural Information Processing Systems 2023, NeurIPS 2023}.

\bibitem[{Svenstrup, Hansen, and Winther(2017)}]{DBLP:conf/nips/SvenstrupHW17}
Svenstrup, D.; Hansen, J.~M.; and Winther, O. 2017.
\newblock Hash Embeddings for Efficient Word Representations.
\newblock In \emph{Advances in Neural Information Processing Systems 30: Annual Conference on Neural Information Processing Systems 2017}, 4928--4936.

\bibitem[{Voigt and Von~dem Bussche(2017)}]{voigt2017eu}
Voigt, P.; and Von~dem Bussche, A. 2017.
\newblock The EU General Data Protection Regulation (GDPR).
\newblock \emph{A Practical Guide, 1st Ed., Cham: Springer International Publishing}, 10: 10--5555.

\bibitem[{Wang et~al.(2025)Wang, Bai, Huang, Li, Wang, and Li}]{DBLP:conf/aaai/WangBHLWL25}
Wang, Z.; Bai, H.; Huang, W.; Li, D.; Wang, J.; and Li, B. 2025.
\newblock Federated Recommendation with Explicitly Encoding Item Bias.
\newblock In \emph{AAAI-25, Sponsored by the Association for the Advancement of Artificial Intelligence}, 12792--12800.

\bibitem[{Wei et~al.(2020)Wei, Li, Ding, Ma, Yang, Farokhi, Jin, Quek, and Poor}]{DBLP:journals/tifs/WeiLDMYFJQP20}
Wei, K.; Li, J.; Ding, M.; Ma, C.; Yang, H.~H.; Farokhi, F.; Jin, S.; Quek, T. Q.~S.; and Poor, H.~V. 2020.
\newblock Federated Learning With Differential Privacy: Algorithms and Performance Analysis.
\newblock \emph{{IEEE} Trans. Inf. Forensics Secur.}, 15: 3454--3469.

\bibitem[{Wu et~al.(2022)Wu, Wu, Lyu, Qi, Huang, and Xie}]{wu2022federated}
Wu, C.; Wu, F.; Lyu, L.; Qi, T.; Huang, Y.; and Xie, X. 2022.
\newblock A Federated Graph Neural Network Framework for Privacy-preserving Personalization.
\newblock \emph{Nature Communications}, 13: 3091.

\bibitem[{Xu et~al.(2024)Xu, Chen, Li, Yang, Wang, and Ngai}]{xu2024aligngroup}
Xu, J.; Chen, Z.; Li, J.; Yang, S.; Wang, H.; and Ngai, E.~C. 2024.
\newblock Aligngroup: Learning and Aligning Group Consensus with Member Preferences for Group Recommendation.
\newblock In \emph{Proceedings of the 33rd ACM International Conference on Information and Knowledge Management}, 2682--2691.

\bibitem[{Yang et~al.(2020)Yang, Tan, Zheng, Chen, and Yang}]{DBLP:series/lncs/YangTZCY20}
Yang, L.; Tan, B.; Zheng, V.~W.; Chen, K.; and Yang, Q. 2020.
\newblock Federated Recommendation Systems.
\newblock In \emph{Federated Learning - Privacy and Incentive}, volume 12500, 225--239.

\bibitem[{Yang et~al.(2024{\natexlab{a}})Yang, Chang, Lai, Yang, Li, Lu, Wang, Yin, and Min}]{yang2024hyperbolic}
Yang, X.; Chang, H.; Lai, Z.; Yang, J.; Li, X.; Lu, Y.; Wang, S.; Yin, D.; and Min, E. 2024{\natexlab{a}}.
\newblock Hyperbolic Contrastive Learning for Cross-Domain Recommendation.
\newblock In \emph{Proceedings of the 33rd ACM International Conference on Information and Knowledge Management}, 2920--2929.

\bibitem[{Yang et~al.(2025{\natexlab{a}})Yang, Jing, Zhang, Wang, Niu, Wang, Lu, Wang, Yin, Liu et~al.}]{Darec}
Yang, X.; Jing, H.; Zhang, Z.; Wang, J.; Niu, H.; Wang, S.; Lu, Y.; Wang, J.; Yin, D.; Liu, X.; et~al. 2025{\natexlab{a}}.
\newblock Darec: A Disentangled Alignment Framework for Large Language Model and Recommender System.
\newblock In \emph{2025 IEEE 41st International Conference on Data Engineering (ICDE)}, 904--917. IEEE.

\bibitem[{Yang et~al.(2024{\natexlab{b}})Yang, Li, Chang, Yang, Yang, Tao, Chang, Shigeno, Wang, Yin et~al.}]{yang2024hgformer}
Yang, X.; Li, X.; Chang, H.; Yang, J.; Yang, X.; Tao, S.; Chang, N.; Shigeno, M.; Wang, J.; Yin, D.; et~al. 2024{\natexlab{b}}.
\newblock Hgformer: Hyperbolic Graph Transformer for Recommendation.
\newblock \emph{arXiv preprint arXiv:2502.15693}.

\bibitem[{Yang et~al.(2025{\natexlab{b}})Yang, Wang, Chen, Fan, Zhao, Zhu, Liu, and Lian}]{TTTRec}
Yang, X.; Wang, Y.; Chen, J.; Fan, W.; Zhao, X.; Zhu, E.; Liu, X.; and Lian, D. 2025{\natexlab{b}}.
\newblock Dual Test-Time Training for Out-of-Distribution Recommender System.
\newblock \emph{IEEE Transactions on Knowledge and Data Engineering}, 37(6): 3312--3326.

\bibitem[{Yuan et~al.(2025)Yuan, Zhang, Sheng, Yao, Zhou, He, and Wang}]{yuan2025pkgrec}
Yuan, H.; Zhang, Y.; Sheng, Q.~Z.; Yao, L.; Zhou, Y.; He, X.; and Wang, Z. 2025.
\newblock PKGRec: Personal Knowledge Graph Construction and Mining for Federated Recommendation Enhancement.
\newblock In \emph{Proceedings of the 34th ACM International Conference on Information and Knowledge Management}, 3973--3982.

\bibitem[{Zhang et~al.(2023{\natexlab{a}})Zhang, Long, Zhou, Yan, Zhang, Zhang, and Yang}]{DBLP:conf/ijcai/ZhangL0YZZY23}
Zhang, C.; Long, G.; Zhou, T.; Yan, P.; Zhang, Z.; Zhang, C.; and Yang, B. 2023{\natexlab{a}}.
\newblock Dual Personalization on Federated Recommendation.
\newblock In \emph{Proceedings of the Thirty-Second International Joint Conference on Artificial Intelligence, {IJCAI} 2023}, 4558--4566.

\bibitem[{Zhang et~al.(2023{\natexlab{b}})Zhang, Luo, Wu, He, and Li}]{DBLP:journals/tois/ZhangLWHL23}
Zhang, H.; Luo, F.; Wu, J.; He, X.; and Li, Y. 2023{\natexlab{b}}.
\newblock LightFR: Lightweight Federated Recommendation with Privacy-preserving Matrix Factorization.
\newblock \emph{{ACM} Trans. Inf. Syst.}, 41: 90:1--90:28.

\bibitem[{Zhang et~al.(2023{\natexlab{c}})Zhang, Zhao, Miao, Shao, Liu, Yang, and Cui}]{DBLP:journals/pvldb/ZhangZMSLYC23}
Zhang, H.; Zhao, P.; Miao, X.; Shao, Y.; Liu, Z.; Yang, T.; and Cui, B. 2023{\natexlab{c}}.
\newblock Experimental Analysis of Large-scale Learnable Vector Storage Compression.
\newblock \emph{Proc. {VLDB} Endow.}, 17: 808--822.

\bibitem[{Zhang et~al.(2025)Zhang, Zhou, Shen, and Li}]{DBLP:journals/tnn/ZhangZSL25}
Zhang, H.; Zhou, X.; Shen, Z.; and Li, Y. 2025.
\newblock PrivFR: Privacy-Enhanced Federated Recommendation With Shared Hash Embedding.
\newblock \emph{{IEEE} Trans. Neural Networks Learn. Syst.}, 36: 32--46.

\bibitem[{Zhao et~al.(2025)Zhao, Bai, Sun, and Yan}]{DBLP:journals/iotj/ZhaoBSY25}
Zhao, X.; Bai, X.; Sun, G.; and Yan, Z. 2025.
\newblock Asynchronous Federated Learning With Local Differential Privacy for Privacy-Enhanced Recommender Systems.
\newblock \emph{{IEEE} Internet Things J.}, 12: 7915--7929.

\bibitem[{Zhu et~al.(2021)Zhu, Liu, Yang, Zhang, and He}]{DBLP:conf/cikm/ZhuLYZH21}
Zhu, J.; Liu, J.; Yang, S.; Zhang, Q.; and He, X. 2021.
\newblock Open Benchmarking for Click-Through Rate Prediction.
\newblock In \emph{{CIKM} '21: The 30th {ACM} International Conference on Information and Knowledge Management}, 2759--2769.

\end{thebibliography}

\clearpage

\end{document}


\appendix
\section{More Implementation Details}
For all FR backbones, the embedding dimension $k$ for both users and items is set to $32$. 
In FedNCF, the MLP architecture is $[64, 128, 64]$ with ReLU activation and a dropout rate of $0.5$. 
In FedPerGNN, the number of message-passing steps is set to $1$.
In PFedRec, the MLP architecture is $[32, 64, 32]$ with ReLU activation and a dropout rate of $0.5$.
The learning rate is selected $(1e-2, 1e-3, 1e-4)$. 

\bigskip